# The Ariadne's Clew Algorithm

**Emmanuel Mazer**                                               EMMANUEL.MAZER@IMAG.FR
*Laboratoire GRAVIR,*
*INRIA 665 Avenue de L'Europe*
*38330 Montbonnot, France*

**Juan Manuel Ahuactzin**                                         JMAL@MAIL.UDLAP.MX
*Depto. de Ing. en Sistemas Computationales*
*Unversidad de las Americas*
*Puebla, Cholula, Puebla 72820, Mexico*

**Pierre Bessière**                                              PIERRE.BESSIERE@IMAG.FR
*Laboratoire LEIBNIZ,*
*46 Avenue Felix Viallet*
*38000 Grenoble, France*

## Abstract

We present a new approach to path planning, called the "Ariadne's clew algorithm". It is designed to find paths in high-dimensional continuous spaces and applies to robots with many degrees of freedom in static, as well as dynamic environments — ones where obstacles may move. The Ariadne's clew algorithm comprises two sub-algorithms, called SEARCH and EXPLORE, applied in an interleaved manner. EXPLORE builds a representation of the accessible space while SEARCH looks for the target. Both are posed as optimization problems. We describe a real implementation of the algorithm to plan paths for a six degrees of freedom arm in a dynamic environment where another six degrees of freedom arm is used as a moving obstacle. Experimental results show that a path is found in about one second without any pre-processing.

## 1. Introduction

The path planning problem is of major interest for industrial robotics. A basic version of this problem consists of finding a sequence of motions for a robot from a start configuration to a given goal configuration while avoiding collisions with any obstacles in the environment.

A simple version of the problem, that of planning the motion of a point robot among 3-dimensional polyhedral obstacles, has been proved to be NP-complete (Canny, 1988). Generally speaking, the complexity of the problem is exponential in the number of degrees of freedom (DOF) of the robot, and polynomial in the number of obstacles in the environment. Consequently, finding a path for a robot with many DOF (more than five) in an environment with several obstacles is, indeed, a very difficult problem. Unfortunately, many realistic industrial problems deal with robots of at least six DOF and hundreds of obstacles. Even worse, often the environment is dynamic in the sense that some of the obstacles may move, thereby further requiring that new paths be found in very short computing times.

In this paper, we present a new approach to path planning, called the "Ariadne's clew algorithm [1]". The approach is completely general and applies to a broad range of path planning problems. However, it is particularly designed to find paths for robots with many DOF in dynamic environments.

The ultimate goal of a planner is to find a path from the initial position to the target. However, while searching for this path, the algorithm may consider collecting information about the free space and about the set of possible paths that lie in that free space. The Ariadne's clew algorithm tries to do both at the same time: a sub-algorithm called EXPLORE collects information about the free space with increasingly fine resolution, while, in parallel, an algorithm called SEARCH opportunistically checks whether the target can be reached.

The EXPLORE algorithm works by placing landmarks in the searched space in such a way that a path from the initial position to any landmark is known. In order to learn as much as possible about the free space, the EXPLORE algorithm tries to spread the landmarks uniformly all over the space. To do this, it places the landmarks as far as possible from one another. For each new landmark produced by the EXPLORE algorithm, the SEARCH algorithm checks (with a local method) whether the target can be reached from that landmark. Both the EXPLORE and SEARCH algorithms are posed as optimization problems.

The Ariadne's clew algorithm is *efficient* and *general*:

1. The algorithm is efficient in two senses:

    (a) Experiments show that the algorithm is able to solve path planning problems fast enough to move a six DOF arm in a realistic and dynamic environment where another six DOF robot is used as a moving obstacle.

    (b) It is well suited for parallel implementation and shows significant speed-up when the number of processors increases.

2. The algorithm is general in two senses:

    (a) It may be used for a wide range of applications in robotics with little additional effort to adapt it.

    (b) Finally, the algorithm is general in that it may be adapted for a large range of search problems in continuous spaces that arise in fields that are not related to robotics.

The paper is organized as follows. Section 2 presents the path planning problem and discusses related work. Section 3 presents the principle of the Ariadne's clew algorithm. Section 4 describes the application of the algorithm to a six DOF arm in a dynamic environment. Finally, Section 5 concludes the paper with a discussion of the contributions of our approach, the main difficulties involved, and possible improvements of our method.

---

1. According to Greek legend, Ariadne, the daughter of Minos, King of Crete, helped Theseus kill the Minotaur, a monster living in the Labyrinth, a huge maze built by Daedalus. The main difficulty faced by Theseus was to find his way through the Labyrinth. Ariadne brilliantly solved the problem by giving him a thread (or a clew) that he could unwind in order to find his path back.

## 2. The Path Planning Problem

Many versions of the path planning problem exist. An exhaustive classification of these problems and of the methods developed to solve them can be found in a survey by Hwang and Ahuja (1992). We choose to illustrate our discussion with a particular case. A robot arm is placed among a set of obstacles. Given an initial and a final position of the robot arm, the problem is to find a set of motions that will lead the robot to move between the two positions without colliding with the obstacles.

To drive the robot amidst the obstacles, early methods (Brooks, 1983) directly used the 3D CAD models of the robot and of the obstacles to find a solution, i.e., they considered the "operational 3D space". In this space, the path planning problem consists of finding the movements of a complex 3D structure (the robot) in a cluttered 3D space.

A major advance was to express the problem in another space known as the configuration space, denoted by $\mathcal{C}$ (Lozano-Pérez, 1987). In this space, the position (or configuration) of a robot is completely determined by a single point having $n$ independent parameters as coordinates. The positions that are not physically legal (because of a collision) are represented by particular regions of $\mathcal{C}$, and are called $\mathcal{C}$-*obstacles*. In the configuration space, the path planning problem consists of finding a continuous curve (representing a path for a single geometrical point) that (i) connects the points representing the initial and the final configuration of the robot, and (ii) does not intersect any $\mathcal{C}$-*obstacles*. This method trades a simplification of the path planning problem (it searches a path for a single point) against a higher-dimensional search space (the dimension of $\mathcal{C}$ is the number DOF of the robot) and against more complex shapes of obstacles (very simple physical obstacles may result in very complex $\mathcal{C}$-*obstacles*).

For example, let us consider the planar arm of Figure 1. Its position among the obstacles is totally known once the values of the angles between its links $(q_0, q_1)$ are known. Thus, for each pair $(q_0, q_1)$, it is possible to decide whether the robot collides with the surrounding obstacles. This is what we did in Figure 2 to represent the mapping between the physical obstacles in the operational space and the $\mathcal{C}$-*obstacles*. Now, by moving a point along the curve joining $\hat{q}_a$ and $\hat{q}_f$ one will also define a collision-free motion for the planar arm between the corresponding positions $P(\hat{q}_a)$ and $P(\hat{q}_f)$ in the operational space. This curve is one solution to this particular path planning problem.

A recent trend in the field is to consider the "trajectory space" (Ferbach, 1996) where a whole path is represented by a single point. The coordinates of this point are the values of the parameters defining the successive movements of the robot. For instance, the list of successive commands sent to the robot controller indeed encode an entire path of the robot. In this space, the path planning problem is reduced to the search for a single point. Once again, we trade a simplification of the path planning problem (searching for a point) against a higher dimension of the search space (the dimension of the trajectory space is the number of parameters needed to specify completely a whole path). For example, in Figure 2, the path between $\hat{q}_b$ and $\hat{q}_d$ can be represented by a point in a seven-dimensional space simply by considering the length of its seven segments.

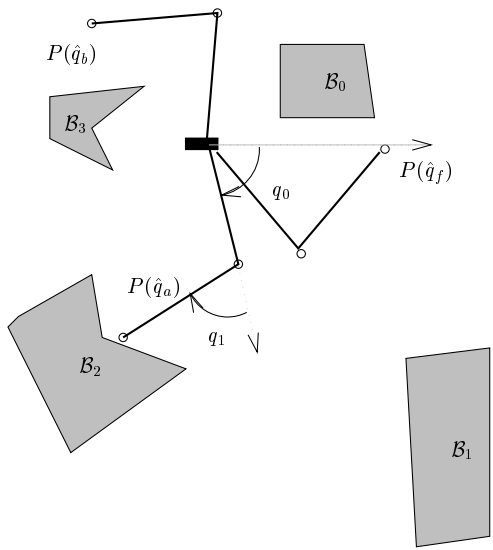

Figure 1: A two DOF arm placed among obstacles in the operational space

## 2.1 Global Approaches

Global approaches are classically divided into two main classes: (i) retraction methods, and (ii) decomposition methods. In the retraction methods, one tries to reduce the dimension of the initial problem by recursively considering sub-manifolds of the configuration space. In the decomposition methods, one tries to characterize the regions of the configuration space that are free of obstacles. Both methods end up with a classical graph search over a discrete space. In principle, these methods are *complete* because they will find a path if one exists and will still terminate in a finite time if a path does not exist. Unfortunately, computing the retraction or the decomposition graph is an NP-complete problem: the complexity of this task grows exponentially as the number of DOF increases (Canny, 1988). Consequently, these planners are used only for robots having a limited number (three or four) of DOF. In addition, they are slow and can only be used off-line: the planner is invoked with a model of the environment, it produces a plan that is passed to the robot controller which, in turn, executes it. In general, the time necessary to achieve this is not short enough to allow the robot to move in a dynamic environment.

## 2.2 Path Planning with Local Planners

One way to combat the complexity of the problem is to trade completeness against performance. To do this, the local planners are guided by the gradient of a cost function (usually the Euclidean distance to the goal) and take into account the constraints introduced by the obstacles to avoid them (Faverjon & Tournassoud, 1987). Since the path planning problem is NP-complete, knowing the cost function, it is always possible to design a deceptive environment where the method will get trapped in a local minimum. However, these methods are useful in many industrial applications because they can deal with complex robots and environment models having thousands of faces, that are often too time-consuming for global methods.

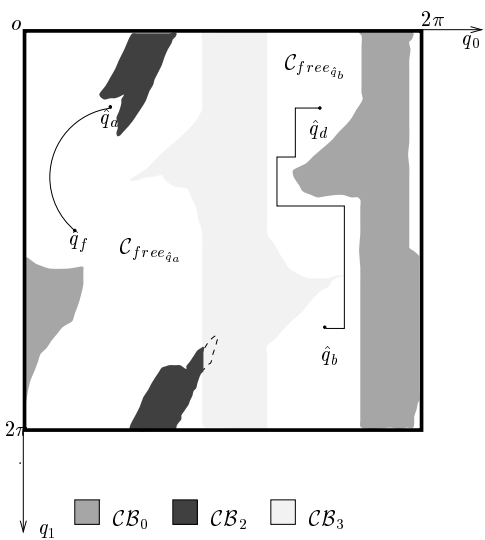

Figure 2: The configuration space corresponding to Figure 1. Note: (1) $\mathcal{C}$ is a torus, (2) it is divided into two regions $\mathcal{C}_{free_{\hat{q}_a}}$ and $\mathcal{C}_{free_{\hat{q}_b}}$ that cannot be connected by a continuous path, and (3) there is not a $\mathcal{C}$-obstacle for $\mathcal{B}_1$ because it does not interfere with the arm.

### 2.3 Path Planning with Randomized Techniques and Landmarks

The stochastic or random approach was first introduced by Barraquand and Latombe (1990), and later used by Overmars (1992), and more recently by Kavraki (1996). The main idea behind these algorithms is to build a graph in the configuration space. The graph is obtained incrementally as follows: a local planner is used to try to reach the goal. Should the motion stop at a local minimum, a new node (or landmark) is created by generating a random motion starting from that local minimum. The method iterates these two steps until the goal configuration has been reached from one of these intermediary positions by a gradient descent motion. These algorithms work with a discretized representation of the configuration space. They are known to be *probabilistically complete* because the probability of terminating with a solution (a path has been found or no path exists) converges to one as the allowed time increase towards infinity. As in the previous section, it is also possible to design simple deceptive environments that will make this kind of algorithm slower than a pure random approach. However, they have been tested for robots with a high number of DOF and they have been shown to work quickly in relatively complex and natural environments.

Other methods using landmarks have been devised. For example, SANDROS, introduced by Chen and Hwang(1992), makes use of landmarks to approximate the free space. This approach is similar to the "hierarchical planning" approach used in AI: should the method fail to reach a goal, new subgoals are generated until the problem is easy enough to be solved. In their approach, first a local planner is used to reach the final position: should the local planner fail, the configuration space is divided into two subspaces, one containing the goal and the other a new sub-goal. The problem is therefore divided into two sub-problems: (i) going from the initial position to the subgoal, and (ii) going from the subgoal to the final position. SANDROS has been shown to be particularly well adapted to find paths for manipulators. It has been implemented and tested for planning paths for Puma and Adept robots.

### 2.4 Path Planning in the Trajectory Space

The previous methods were essentially based on the configuration space: the retraction, the decomposition, or the optimization is made in this space. An alternative is to consider the "trajectory space". For example, in his method VDP, Ferbach (1996) starts by considering the straight line segment joining the initial and the final configuration in $\mathcal{C}$. This path is progressively modified in such a manner that the forbidden regions it crosses are reduced. At each iteration, a sub-manifold of $\mathcal{C}$ containing the current path is randomly generated. It is then discretized and explored using a dynamic programming method that uses the length across the forbidden region as the cost function in order to minimize. The search results in a new trajectory whose intersection with the forbidden regions is smaller than the original trajectory. The process is repeated until an admissible trajectory is found. As in the previous sections, it is also possible to design simple deceptive environments that will make this kind of algorithm slower than a pure random approach.

The work of Lin, Xiao, and Michalewicz (1994) is similar to our approach. As in an early version of our algorithm (Ahuactzin, Mazer, Bessière, & Talbi, 1992), genetic algorithms are used to carry out optimization in the trajectory space. Trajectories are parameterized using the coordinates of intermediary via-points. An evolutionary algorithm is used to optimize a cost function based on the length of the trajectory and the forbidden region crossed. The standard operators of the genetic algorithms have been modified and later extended to produce a large variety of paths (Xiao, Michalewicz, & Zhang, 1996). The number of intermediary via-points is fixed and chosen using an heuristic. Given this number, nothing prevents to design a deceptive problem which solution will require more intermediary points, leading the algorithm to fail while one solution exists.

## 3. Principle of the Ariadne's Clew Algorithm

As we have seen in the previous section, the computation of the configuration space $\mathcal{C}$ is a very time-consuming task. The main idea behind the Ariadne's clew algorithm is to avoid this computation. In order to do this, the algorithm searches directly for a feasible path in the trajectory space. The configuration space $\mathcal{C}$ is never explicitly computed.

As will be shown, in the trajectory space, path planning may be seen as an optimization problem and solved as such by an algorithm called SEARCH. It is possible to build an approximation of free space by another algorithm called EXPLORE that is also posed as an optimization problem. The Ariadne's clew algorithm is the result of the interleaved execution of SEARCH and EXPLORE.

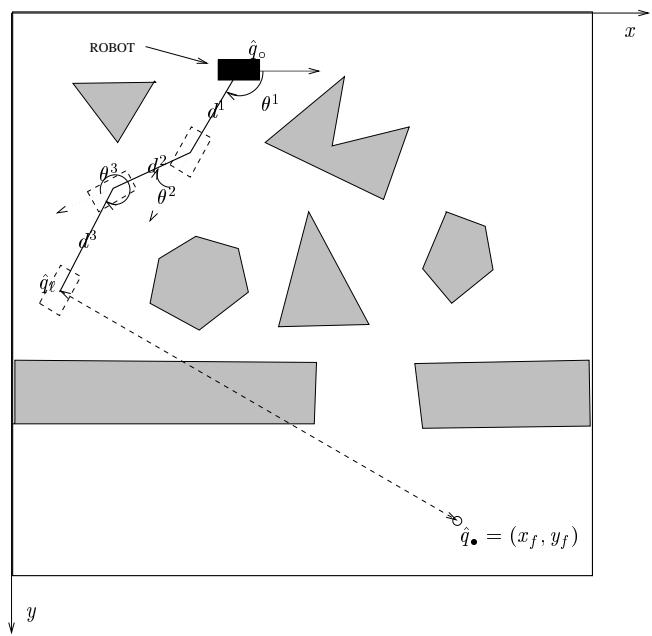

Figure 3: A parameterized trajectory $(\theta^1, d^1, \theta^2, d^2, ...\theta^l, d^l)$ and a starting point $\hat{q}_0$ implicitly define a path (in the operational space) for a holonomic mobile robot.

## 3.1 Path Planning as an Optimization Problem: SEARCH

Given a robot with $k$ DOF, a trajectory of length $l$ may be parameterized as a sequence of $n = k * l$ successive movements. A starting point $\hat{q}_\circ$ along with such a parameterized trajectory implicitly define a path and a final configuration $\hat{q}_l$ in the configuration space. For example, for a holonomic mobile robot the trajectory $(\theta^1, d^1, \theta^2, d^2, ...\theta^l, d^l)$ can be interpreted as making a $\theta^1$ degree turn, moving straight $d^1$, making a $\theta^2$ degree turn and so on. Given the starting configuration $\hat{q}_\circ$, this trajectory leads to the final configuration $\hat{q}_l$ (see Figure 3).

Given a distance function $d$ on the configuration space, if we find a trajectory such that it does not collide with any obstacles and such that the distance between $\hat{q}_l$ and the goal $\hat{q}_\bullet$ is zero, then we have a solution to our path planning problem. Therefore, the path planning problem may be seen as a minimization problem where:

1. The search space is a space of suitably parameterized trajectories, the trajectory space.

2. The function to minimize is $d(\hat{q}_l, \hat{q}_\bullet)$ if the path is collision-free, and $d(\hat{q}_i, \hat{q}_\bullet)$ otherwise ($\hat{q}_i$ being the first collision point).[2]

---

2. Another possible choice would be to give the $+\infty$ value to the distance when a collision occurs. However, this is less informative than the chosen function because the first part of a colliding path could be a good start toward the goal and should not be discarded. Note that the cost function does not include any optimality criteria such as the length of the trajectory or the amount of energy used.

The algorithm SEARCH, based on this very simple technique and a randomized optimization method, is already able to solve quite complex problems of robot motion planning. For example, Figure 4 represents the two paths found for the holonomic mobile robot. Each path was computed on a standard workstation (SPARC 5) in less than 0.5 second without using any pre-computation of the configuration space. Thus, it is possible, albeit slowly, to get a planner that can be used in a dynamic environment (where the obstacles may move) by "dropping" a new world into the system every 0.5 second. SEARCH is very efficient but it is not complete, since it may fail to find a path even if one exists for two different reasons:

1. Due to the optimization-based formulation, SEARCH can get trapped by local minima of the objective function, which in turn may place the robot far away from the goal (see Figure 5).

2. The length $l$ of the trajectories considered may be too short to reach all the accessible regions of the configuration space.

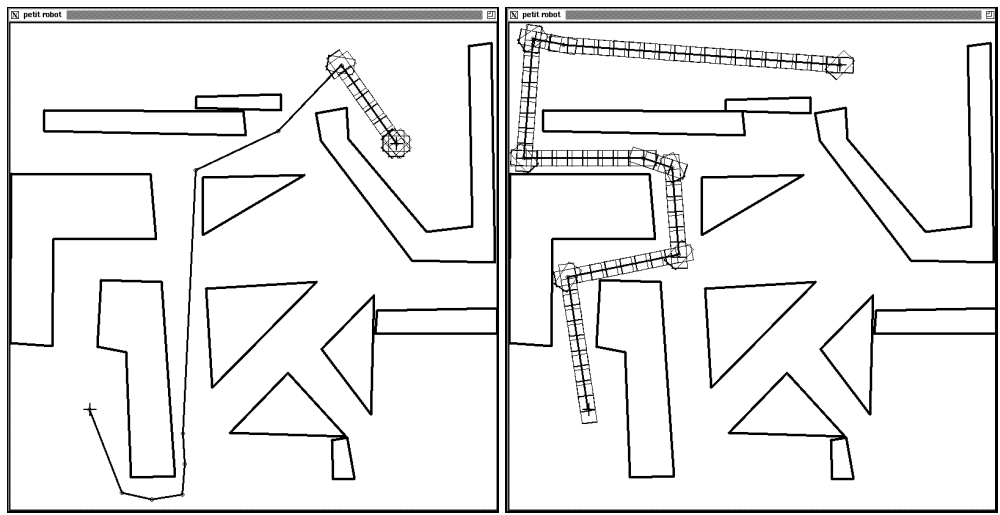

Figure 4: Reactive replanning in a changing environment

### 3.2 Exploring as an Optimization Problem: EXPLORE

In order to build a complete planner, we propose a second algorithm called EXPLORE. While the purpose of SEARCH was to look directly for a path from $\hat{q}_\circ$ to $\hat{q}_\bullet$, the purpose of EXPLORE is to compute an approximation of the region of the configuration space accessible from $\hat{q}_\circ$.

The EXPLORE algorithm builds an approximation of the accessible space by placing landmarks in the configuration space $\mathcal{C}$ in such a way that a path from the initial position $\hat{q}_\circ$ to any landmark is known. In order to learn as much as possible about the free space, the EXPLORE algorithm tries to spread the landmarks uniformly over the space (see Figure 6). To do this, it tries to put the landmarks as far as possible from one another by maximizing the distances between them.

Therefore, EXPLORE may be seen as a maximization problem where:

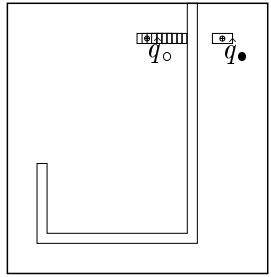

Figure 5: A problem leading to a local minimum. In such a case, a solution path has first to move away from the goal. The goal's "attraction" based on the minimization of the Euclidean distance prevents SEARCH from finding such a path.

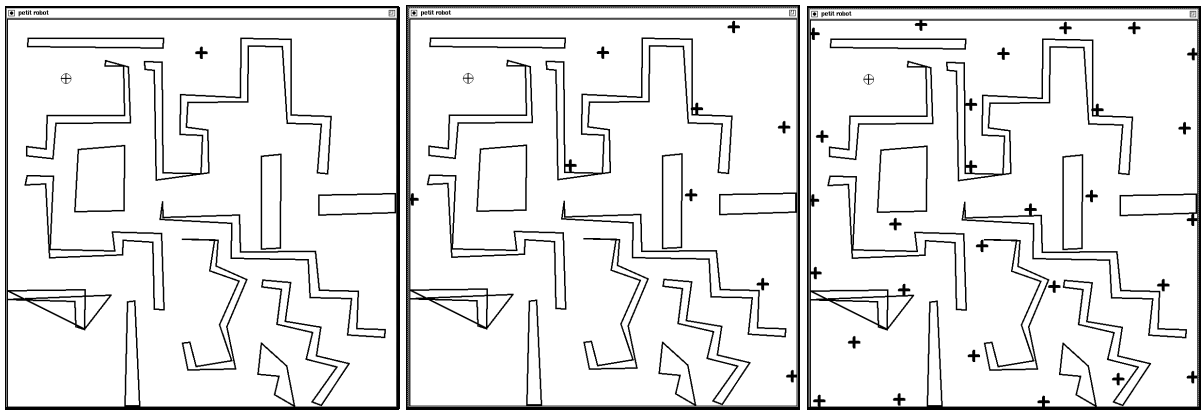

Figure 6: The first picture represents the initial position and the first landmark. The subsequent landmarks are then uniformly spread over the search space while the method keeps track of all paths joining the landmarks to the initial position. The algorithm is named after Ariadne because by placing landmarks, EXPLORE unwinds as if it were using a thread as Theseus did.

1. The search space is the set of all paths starting from one of the previously placed landmarks.

2. The function to maximize is $d(\hat{q}_l, \Lambda)$, where $\Lambda$ is the set of landmarks already placed.

### 3.3 The Ariadne's Clew Algorithm: EXPLORE + SEARCH

In order to have a planner that is both complete and efficient, we combined the two previous algorithms SEARCH and EXPLORE to obtain the Ariadne's clew algorithm.

The principle of the Ariadne's clew algorithm is very simple:

1. Use the SEARCH algorithm to find whether a "simple" path exists between $\hat{q}_\circ$ and $\hat{q}_\bullet$.

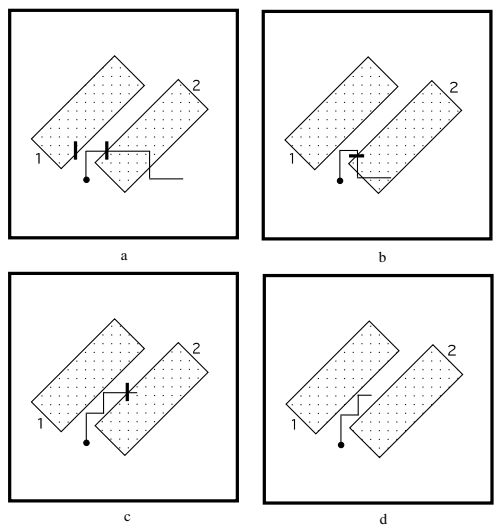

Figure 7: Bouncing against $\mathcal{C}$-obstacles. Figure (a) presents the original path in the configuration space. Figure (b) shows the same path after two bounces along the second segment on obstacle 2 and on obstacle 1. Figure (c) is the result obtained after a bounce of segment 3 against obstacle 2. Finally, Figure (d) presents a valid path obtained after a final bounce of segment 4 against obstacle 2.

2. If no "simple" path is found by step 1, then continue until a path is found.

   (a) Use EXPLORE to generate a new landmark.
   (b) Use SEARCH to look for a "simple" path from that landmark to $\hat{q}_\bullet$.

The Ariadne's clew algorithm will find a path if one exists. In an overwhelming number of cases, just a few landmarks are necessary for the Ariadne's clew algorithm to reach the target and stop.

### 3.4 A Major Improvement: Bouncing on $\mathcal{C}$-Obstacles

A typical difficulty for a path planning algorithm is to find a collision-free path through a small corridor in the configuration space. This is also the case for the basic version of the Ariadne's clew algorithm, presented above. The problem is that very few trajectories encode such paths and therefore they are very difficult to find. Most trajectories collide with the obstacles. We propose a very simple idea to deal with this problem: going backwards at each collision point. If, for a given trajectory, a collision is detected along the corresponding path, then we simply consider transforming that trajectory so that it encodes a new path, one that is found by bouncing off the obstacle at the collision point (see Figure 7). Note that this construction is applied recursively until the entire trajectory corresponds to a collision-free path.

THE ARIADNE'S CLEW ALGORITHM

Using this technique, all trajectories are so transformed that they encode valid paths. This improved version of the Ariadne's clew algorithm no longer cares about obstacles. From the point of view of a search in the trajectory space, it is as if the obstacles have simply vanished. This method is especially efficient for narrow corridors in the configuration space. Without bouncing, the mapping of a corridor in the configuration space to the trajectory space is a set of very few points. With bouncing, every single trajectory going through a part of the corridor is "folded" into the corridor (see Figure 7). The resultant mapping of the corridor in the trajectory space is consequently a much larger set of points, and therefore it is much easier to find a member of this set. This empirical improvement has a major practical impact because it makes the proposed algorithm faster (fifteen times) in the problem considered below.

### 3.5 The Algorithm

We can now give a final version of the Ariadne's clew algorithm. It has three inputs: $\hat{q}_\circ$ (the initial position), $\hat{q}_\bullet$ (the goal position), and $\rho$ (the maximum allowed distance for a path to the $\mathcal{C}$-*obstacles*). It returns a legal path or terminates if no path exists at the given resolution.

```
ALGORITHM_ARIADNE($\hat{q}_\circ, \hat{q}_\bullet, \rho$)
begin
    $i := 1; \hat{\lambda}_1 := \hat{q}_\circ$;
    /* Initialize the set of landmarks with the initial position
    $\Lambda_1 := \{\hat{\lambda}_1\}; \varepsilon_1 = +\infty$;
    do while ($\varepsilon_i > \rho$);
        /* run SEARCH : look for the goal with a local method
        if ($\min_{l \in \mathbb{R}^\ell} d(\hat{q}_\bullet, \hat{q}(l)) == 0$)
            return; /* A path has been found !
        else
        /* run EXPLORE : place a new landmark
            $i := i + 1$;
            $\hat{\lambda}_i := \hat{q} : \sup_{l \in \mathbb{R}^\ell} d(\Lambda_{i-1}, \hat{q}(l))$;
            $\Lambda_i := \Lambda_{i-1} \cup \{\hat{\lambda}_i\}$;
            $\varepsilon_i := d(\Lambda_{i-1}, \hat{\lambda}_i)$;
        endif
    enddo
    $\Lambda = \Lambda_i$;
    $\varepsilon = \varepsilon_i$;
    return($\varepsilon$); /* No path !
end
```

Figure 8: The Ariadne's Clew Algorithm

The algorithm is based on the following optimization problems:

$$EXPLORE : \begin{cases} \sup d(\Lambda_{i-1}, \hat{q}(l)) \\ l \in I\!\!R^\ell \end{cases}$$

$$SEARCH : \begin{cases} \min d(\hat{q}(l), \hat{q}_\bullet) \\ l \in I\!\!R^\ell \end{cases}$$

$\hat{q}(l)$ denotes the extremity of a legal path parameterized with $\ell$ real parameters and starting either from each of the previously placed landmarks (EXPLORE) or from the latest placed landmark (SEARCH).

The algorithm is resolution-complete under the following assumptions:

- "Space filling completeness": The global maximum distance can be found by the optimization algorithm used in EXPLORE; the configuration space is a compact set.

- "$\rho$ completeness": The optimization procedure used in SEARCH always find a complete path (or returns 0) when the starting and the goal positions are located within a ball of radius $\rho$ of the free space.

In practice, the first condition cannot be met with a randomized optimization algorithm in a bounded time, and only local maxima are found. However, the landmarks placed according to the new algorithm are better distributed over the free space than landmarks placed randomly, leading to better performances. The goal of the next section is to justify this claim, experimentally.

## 4. Path Planning for a Six DOF Arm in a Dynamic Environment

In order to demonstrate the feasibility and qualities of the Ariadne's clew algorithm, we have developed a realistic application of the algorithm. We selected a problem where we want to have a path planner for a six DOF robot arm in a dynamic environment where another arm is used as a mobile obstacle. The robot (robot A) is under the control of the Ariadne's clew algorithm. It shares its workspace with a second robot (robot B) that is moving under the control of a random motion generator. The Ariadne's clew algorithm must be able to compute paths for A in "real time" (here, real time means fast enough to ensure that robot A will never collide with robot B).

In order to reach such a level of performance, we chose to implement the Ariadne's clew algorithm on a massively parallel machine (Meganode with 128 T800 Transputers). Furthermore, we selected a genetic algorithm as our optimization technique. The reasons for this choice are:

1. Genetic algorithms are well suited for problems where the search space is huge but where there are many acceptable solutions. This is exactly the case here. The trajectory space is huge but there are, barring exceptional cases, numerous acceptable paths going from $\hat{q}_\circ$ to $\hat{q}_\bullet$ without collision.

2. Genetic algorithms, unlike a number of the other optimization techniques (Bessière, Talbi, Ahuactzin, & Mazer, 1996), are very easy to implement on parallel architectures. We have previously developed a parallel genetic algorithm (PGA) and we have already had significant experience using it (Talbi, 1993).

3. PGA, unlike most parallel programs, shows linear speed-up (when you double the number of processors you reduce the computation time by half) and even super-linear speed-up under certain circumstances (Talbi & Bessière, 1996).

### 4.1 Parallel Genetic Algorithm

Genetic algorithms are stochastic optimization techniques introduced by Holland (1975) twenty years ago. They are used in a large variety of domains including robotics (Ahuactzin et al., 1992; Lawrence, 1991; Falkenauer & Bouffouix, 1991; Falkenauer & Delchambre, 1992; Meygret & Levine, 1992) because they are easy to implement and do not require algebraic expression for the function to be optimized.

#### 4.1.1 PRINCIPLE OF GENETIC ALGORITHM

The goal of the algorithm is to find a point reaching a "good" value of a given function $F$ over a search space $S$. First, a quantization step is defined for S and the search is conducted over a discrete subset, $S_d$ of $S$. $S_d$ contains $2^N$ elements. In practice, the cardinality of $S_d$ can be *extremely* large. For example, in our implementation of EXPLORE, $N = 116$. Thus, a continuous domain is discretized with a given resolution.

During an initialization phase a small subset of $S_d$ is drawn at random. This subset is called a *population*. Each element of this population is coded by a string of $N$ bits.

The genetic algorithm iterates the following four steps until a solution is found.

1. **Evaluation**: Rank the population according to the value of $F$ for each element of $S_d$. Decide if the best element can serve as an acceptable solution; if yes, exit.

2. **Selection**: Use the function $F$ to define a probability distribution over the population. Select a pair of elements randomly according to this probability distribution.

3. **Reproduction**: Produce a new element from each pair using "genetic" operators.

4. **Replacement**: Replace the elements of the starting population by better new elements produced in step 3.

Many genetic operators (Davidor, 1989) are available. However, the more commonly used are the *mutation* and the *cross-over* operators. The mutation operator consists of randomly flipping some bits of an element of the population. The cross-over operator consists of first randomly choosing a place where to cut the two strings of bits, and then building two new elements from this pair by simply gluing the right and the left parts of the initial pair of strings (see Figure 9).

We use both operators to produce new elements. First, we use the cross-over operator to get an intermediate string. Then, the mutation operator is used on this intermediate string to get the final string.

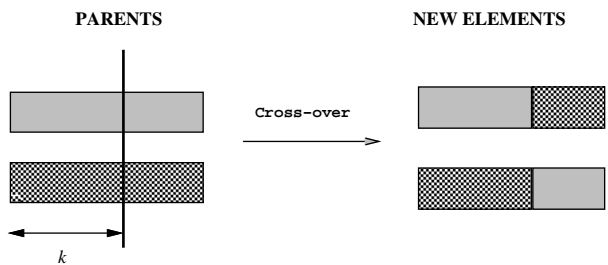

Figure 9: The cross-over operation.

### 4.1.2 Principle of the Parallel Genetic Algorithm (PGA)

There are many parallel versions of genetic algorithms: the standard parallel version (Robertson, 1987), the decomposition version (Tanese, 1987) and the massively parallel version (Talbi, 1993). We chose this last method. The main idea is to allocate one element of the population for each processor so that steps 1, 3, and 4 can be executed in parallel. Furthermore, the selection step (step 2) is carried out locally, in that each individual may mate only with the individuals placed on processors physically connected to it. This ensures that the communication overhead does not increase as the number of processors increases. This is the reason why PGA shows linear speed-up.

The parallel genetic algorithm iterates the following four steps until a solution is found.

1. **Evaluation**: Evaluate *in parallel* all the individuals.

2. **Selection**: Select *in parallel*, among the neighbors, the mate with the best evaluation.

3. **Reproduction**: Reproduce *in parallel* with the chosen mate.

4. **Replacement**: Replace *in parallel* the parents by the offspring.

On the Meganode, we implemented the PGA on a torus of processors where each individual has four neighbors (see Figure 10)

### 4.2 Parallel Evaluation of the Cost Function

The evaluation functions used in SEARCH and EXPLORE are very similar: they both compute the final position of the arm given a Manhattan path of a fixed order. In our implementation, based on experience, we chose to use Manhattan paths of order 2. Order 2 appeared to be a good compromise between the number of landmarks needed (increases as order decreases) and the computing time necessary for the optimization functions (increases as order increases). Since our robot has six DOF, the argument of the cost function in SEARCH is a vector in $R^{12}$: $(\Delta_1^1, \Delta_2^1, ..., \Delta_6^1, \ldots, \Delta_1^2, \ldots, \Delta_6^2)$ and the argument of the cost function used for EXPLORE is a vector in $I\!N \times I\!R^{12}$ : $(i, \Delta_1^1, \Delta_2^1, ..., \Delta_6^1, \ldots, \Delta_1^2, \ldots, \Delta_6^2)$ where $i$ codes the landmark used as a starting point for the path. In both cases the functions are defined only on a bounded subset of $I\!R^{12}$ and $I\!N \times I\!R^{12}$, whose limits are fixed by the mechanical stops of the robot and the maximum number of landmarks. A discretization step is chosen for these two subsets by defining the resolution at which each elementary motion is discretized.

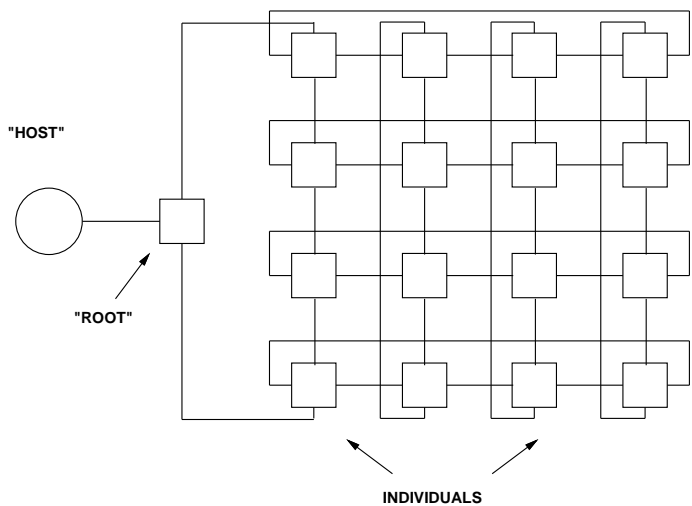

Figure 10: A torus with sixteen processors. One individual is placed on each processor. Each individual has four neighbors.

In our case, each $\Delta_i^j$ is discretized with 9 bits and the number of landmarks is limited to 256. Thus, given a binary string of $116 = 8 + 12 \times 9$ bits, we can convert it into a vector (as an argument) for the cost function of SEARCH, or EXPLORE, respectively.

Manhattan paths are evaluated in a simplified model of the environment. This model is obtained by enclosing each element of the scene into a bounding rectangular box.

The evaluation of a vector is performed as follows:

For each $\Delta_i^j$ in $(\Delta_1^1, \Delta_2^1, ..., \Delta_6^1, \ldots, \Delta_1^2, \ldots, \Delta_6^2)$
    Compute the limits on the motion for joint $i$.
    Compute $\Delta_i^{j'}$ by bouncing on these limits (see Section 3.4).
    Update the position of the robot.

The limits on the motion of joint $i$ are obtained by merging the legal ranges of motion of all the links that move when joint $i$ moves, and all the obstacles. To obtain a legal range of motion between a link and an obstacle, we consider the two enclosing parallelepipeds and express their coordinates in the joint frame. Then, we use a classical method to compute the range (Lozano-Pérez, 1987).

In our parallel implementation, we distributed the geometric computations among several processors. Each processor is dedicated to the computation of a single type of interaction.

### 4.3 Parallel Implementation of the Ariadne's Clew Algorithm

Finally, the Ariadne's clew algorithm is implemented in parallel with three levels of parallelism.

1. Obviously, a first level of parallelization can be obtained by running SEARCH and EXPLORE at the same time on two sets of processors. While SEARCH is checking whether a path exists between the last placed landmark and the goal, EXPLORE is generating the next landmark.

2. The second level of parallelism corresponds to a parallel implementation of both genetic algorithms employed by SEARCH and EXPLORE to treat their respective optimization problems.

3. The third level corresponds to a parallelization of the collision checking function and range computation.

We completed a full implementation of these three levels on a Meganode with 128 T800 transputers. Figure 11 represents our parallel implementation of the Ariadne's clew algorithm and Figure 12 shows how we have embedded this architecture into our experimental setup. A CAD system (ACT) is used to model the scene with the two robots. The robots are under the control of KALI (Hayward, Daneshmend, & Hayati, 1988). First, a simplified geometric model of the scene is downloaded into the memory of the transputers. Then, a Silicon Graphics workstation works as a global controller and loops over the following steps:

1. Generate and execute a legal random motion for robot B.

2. Send the new configuration of robot B to the Meganode as well as the desired final configuration for robot A.

3. Get the planned path for robot A from the Meganode and execute it.

4. Wait for a random time and stop robot A.

5. Go to 1.

This sequence allows us to test our algorithm extensively in real situations by having to deal with many different environments. Of course, the most interesting figure we can obtain from this experiment is the mean time necessary to compute one path given a new environment. For this experimental setup this mean time is 1.421 seconds. Using the same architecture with more up-to-date processors (T9000) would reduce this time by a factor of ten. The same computation on a single processor (SPARC 5) would take three times longer than the current implementation.

*In summary, we have achieved our main goal by proving that it is indeed possible (with the Ariadne's clew algorithm) to plan collision-free paths for a real robot with many DOF in a dynamic realistic environment.*

## 5. Conclusion: Contributions, Difficulties, and Perspectives

As mentioned in the Introduction, the Ariadne's clew algorithm has two main qualities: *efficiency*, and *generality*. Let us, in conclusion, explain and discuss these two qualities.

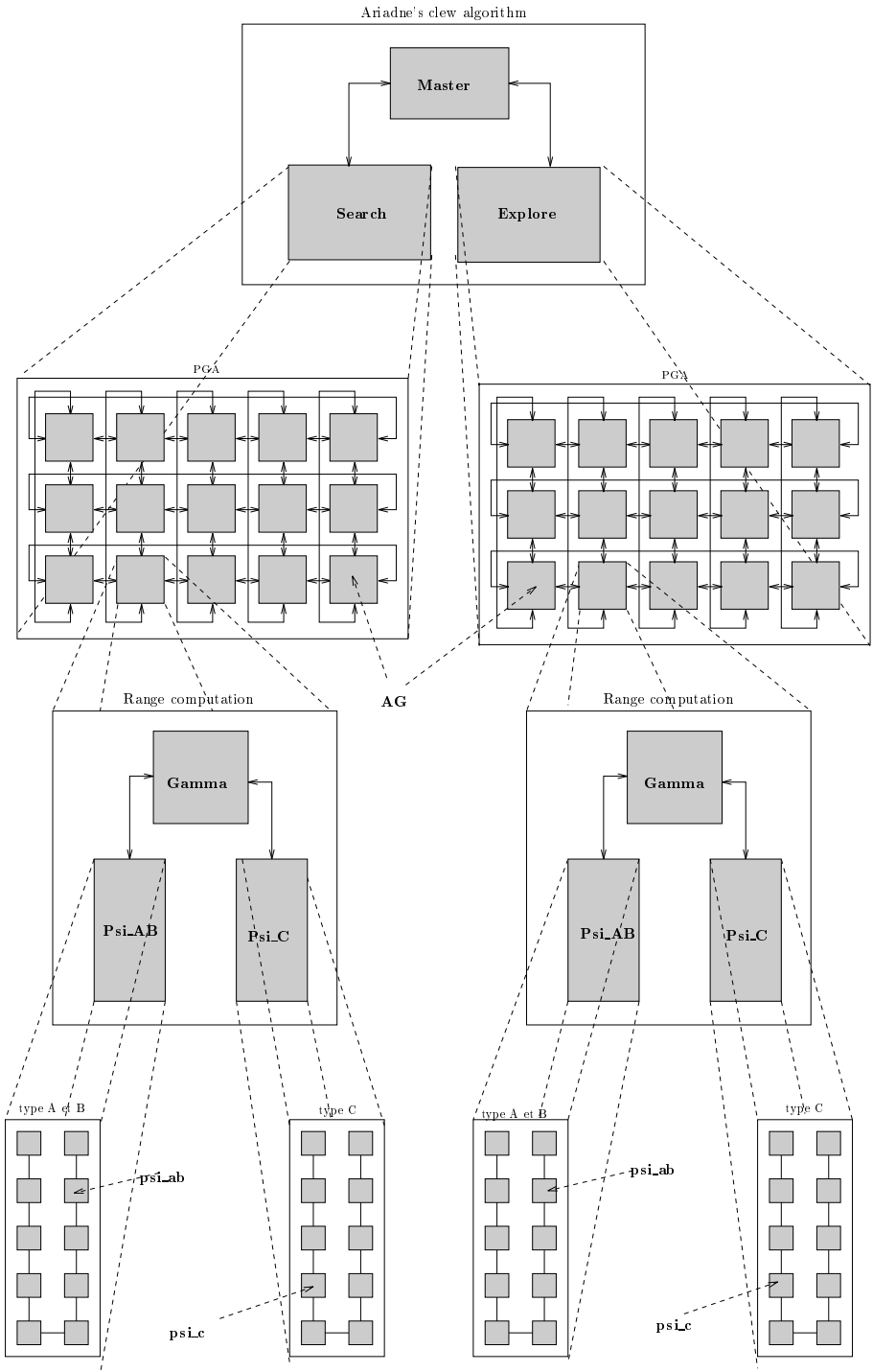

Figure 11: A parallel implementation of the Ariadne's clew algorithm

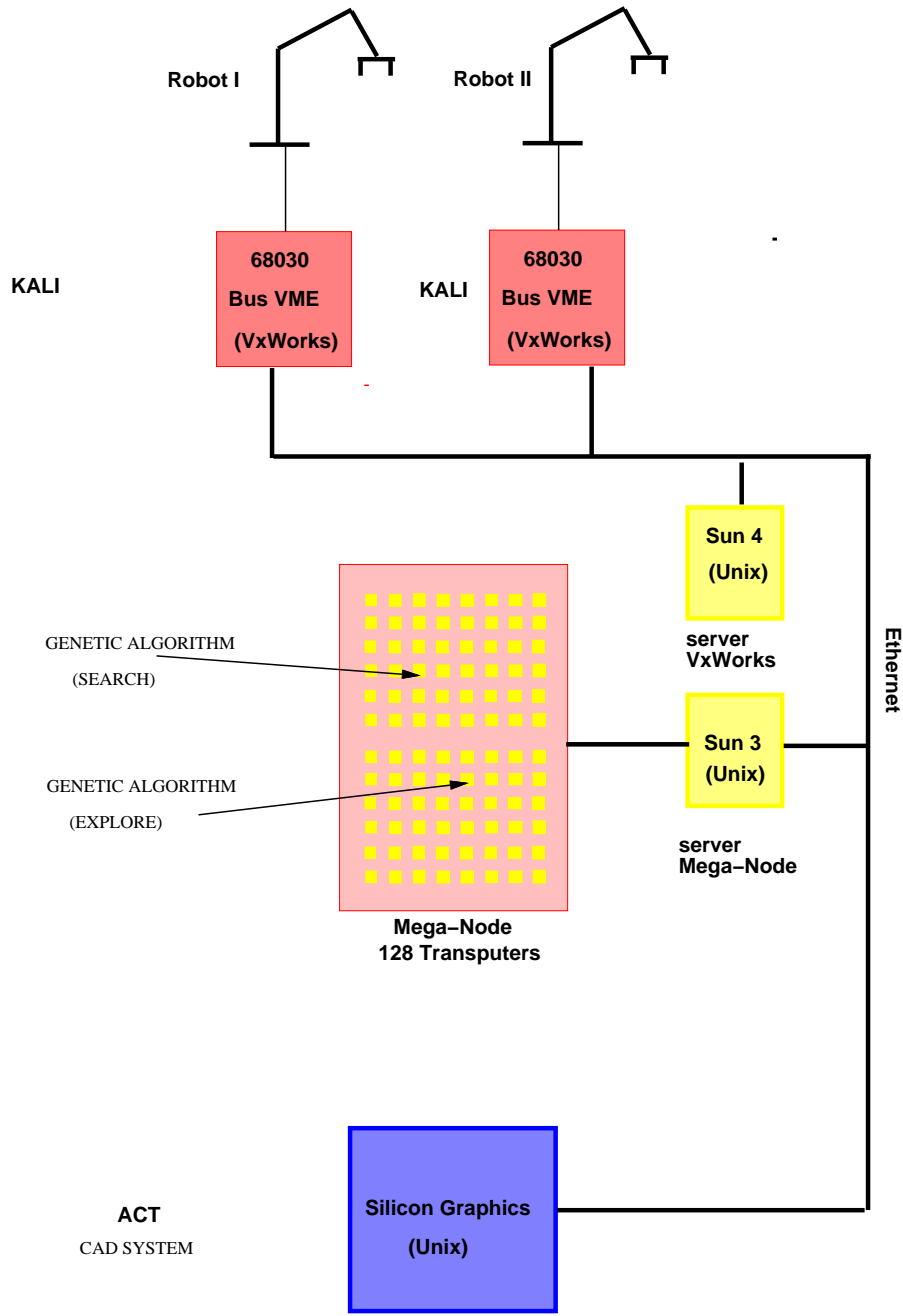

Figure 12: The experimental setup

## 5.1 Performance

Comparing the performance of this kind of algorithm is a very delicate subject. Performance may be a matter of computing time, efforts needed to program, or ease of application to different problems (see Section 5.2). Evaluating the performance in terms of computing time is very difficult for one fundamental and three practical reasons:

THE ARIADNE'S CLEW ALGORITHM

1. The fundamental reason is, once again, the NP-completeness of the path planning problem. As deceptive cases may always be designed, the only performance results one may reasonably present are always specific.

2. The three practical reasons are:

    (a) Obviously, the first requirement for such a comparison is that different algorithms run on the same machines with the same available memory. This may seem simple but it is a main difficulty in our case because our algorithm has been designed to run on rather specific kinds of machines, namely, massively parallel ones. It could also be implemented on non-parallel machines, but then it may lose part of its interest. A fair comparison would be to compare the algorithms on both types of machines. This would imply programming other algorithms in parallel, which is very difficult in practice.

    (b) Many known path planning algorithms first compute the configuration space (or an approximation of it) off-line, and then efficiently solve the path planning problem on-line. As we saw, in order to deal with a dynamic environment, the Ariadne's clew algorithm adopts a completely different approach.

    (c) For practical reasons, many test problems are toy problems (2D, few obstacles, few faces, simulated robots) and the performance results using these kinds of problems are very difficult to generalize to realistic industrial problems (3D, tens of obstacles, hundreds of faces, real robots).

Considering all these reasons, we tested our algorithm by implementing a realistic robotic application to the very end. To achieve this goal, we assembled a complex experimental setup including six different machines (1 MEGANODE, 2 68030, 2 SUN 4, and 1 SILICON GRAPHICS), two mechanical arms, and running seven different cooperative programs (2 KALI, 1 ACT, 2 VXWORKS, 1 PARX, and 1 Ariadne's clew algorithm).

Our challenge was to be able to solve the path planning problem fast enough to drive a real six DOF arm in a dynamic environment. The Ariadne's clew algorithm indeed achieved this goal in our experiments where the environment is composed of five fixed obstacles and a six DOF arm moving independently.

We are not aware of any other methods capable of such performance. To the best of our knowledge, currently implemented planners would take a number of seconds (ten) to place a set of landmarks on a 2D example for a robot with five DOF (Kavraki et al., 1996). Despite the fact that finding a general purpose planning technique for real industrial application is a very difficult problem, we believe that the Ariadne's clew algorithm provides an effective approach to such problems.

The number of range computations for a Manhattan motion of order 1 is $C\frac{k^2+k}{2} * n$ where $n$ is the number of faces, $k$ the number of DOF, and $C$ a constant factor, depending on the number of parts used to model the robot. Obviously, such a number of faces may be a severe difficulty for the implementation of the Ariadne's clew algorithm described so far. To speed up the computation we use a number of geometric filters that reduce the number of pairs of entities to be analyzed.

However, it was possible to follow two research tracks in combination. First, we could use collision checking methods that allow access to the pairs in collision in a logarithmic time (Faverjon & Tournassoud, 1987). Second, we could preserve part of the landmark graph when the environment is changing (McLean & Mazon, 1996).

### 5.2 Generality

The Ariadne's clew algorithm is general in the sense that it may be used for numerous and very different applications in robotics. Basically, the main thing that needs to be changed in the algorithm is the distance $d$ used in the evaluation functions of the two optimization problems.

Several planners have been implemented in this way: a fine motion planner (De la Rosa, Laugier, & Najera, 1996), two motion planners for holonomic and non-holonomic mobile robots (Scheuer & Fraichard, 1997), a reorientation planner for an articulated hand(Gupta, 1995), a planner for grasping and regrasping (Ahuactzin, Gupta, & Mazer, 1998), and a planner for a robotic arm placed in the steam generator of a nuclear plant (McLean & Mazon, 1996). Adapting the algorithm to a new application is, therefore, clearly a very easy task. For instance, the application to path planning for the non-holonomic trailer was developed in three days.

The Ariadne's clew algorithm is also general in the sense that it may be used for any kind of path planning problem in a continuous space, in fields other than robotics. Although it may be sufficient to change the distance function $d$, one may also consider changing the form of the function $d$, or even the nature of the searched spaces. For instance, the concept of obstacles may be reconsidered. Instead of "hard" obstacles, one could replace them by zones of constraints. In that case, the path planning problem does not consist of finding a path without collisions but rather finding a path best satisfying the different constraints. Such a planner has been developed for a naval application where the problem was to find a path for a boat with various constraints on the trajectory. This opens numerous perspectives of applications for applying the Ariadne's clew algorithm in a broader field than pure robotics.

### Acknowledgments

The authors are greatly indebted to Dr. Kamal Gupta from Simon Fraser University who carefully read the paper and suggested valuable corrections that greatly improve the quality of the final paper.

This work has been made possible by: Le Centre National de la Recherche Scientifique (France), Consejo Nacional de Ciencia y Tecnologia (Mexico) and ESPRIT 2, P2528 (EEC).

The Ariadne's clew algorithm